\documentclass[10pt,twocolumn,letterpaper]{article}

\usepackage{cvpr}
\usepackage{times}
\usepackage{epsfig}
\usepackage{graphicx}
\usepackage{amsmath}
\usepackage{amssymb}
\usepackage{color}
\usepackage{ulem}
\usepackage{booktabs}
\usepackage{multirow, makecell}
\usepackage{nicefrac}

\usepackage[pagebackref=true,breaklinks=true,letterpaper=true,colorlinks,bookmarks=false]{hyperref}

\cvprfinalcopy % *** Uncomment this line for the final submission

\def\cvprPaperID{4674} % *** Enter the CVPR Paper ID here

\ifcvprfinal\fi

\begin{document}

%%%%%%%%% TITLE
\title{LaSO: Label-Set Operations networks for multi-label few-shot learning}

\newcommand{\printfnsymbol}[1]{$^*$}
\small{\author{
Amit Alfassy\thanks{The authors have contributed equally to this work}, ~Leonid Karlinsky\printfnsymbol{1}, Amit Aides\printfnsymbol{1}, Joseph Shtok, Sivan Harary, Rogerio Feris\\
IBM Research AI\\
Haifa, Israel \\
\and
Raja Giryes\\
School of Electrical Engineering, Tel-Aviv University\\
Tel-Aviv, Israel\\
\and
Alex M. Bronstein\\
Department of Computer Science, Technion\\
Haifa, Israel\\
}}

\maketitle

\begin{abstract}
Example synthesis is one of the leading methods to tackle the problem of few-shot learning, where only a small number of samples per class are available. However, current synthesis approaches only address the scenario of a single category label per image. In this work, we propose a novel technique for synthesizing samples with multiple labels for the (yet unhandled) multi-label few-shot classification scenario. We propose to combine pairs of given examples in feature space, so that the resulting synthesized feature vectors will correspond to examples whose label sets are obtained through certain set operations on the label sets of the corresponding input pairs.
Thus, our method is capable of producing a sample containing the intersection, union or set-difference of labels present in two input samples. As we show, these set operations generalize to labels unseen during training. This enables performing augmentation on examples of novel categories, thus, facilitating multi-label few-shot classifier learning. We conduct numerous experiments showing promising results for the label-set manipulation capabilities of the proposed approach, both directly (using the classification and retrieval metrics), and in the context of performing data augmentation for multi-label few-shot learning. We propose a benchmark for this new and challenging task and show that our method compares favorably to all the common baselines. Our code will be made available upon acceptance.
\end{abstract}

\section{Introduction}
\label{sec:introduction}

\begin{figure}[ht]
\begin{center}
   %\includegraphics[width=0.5\textwidth]{figures/setOps_fig.png}
   \includegraphics[width=0.5\textwidth]{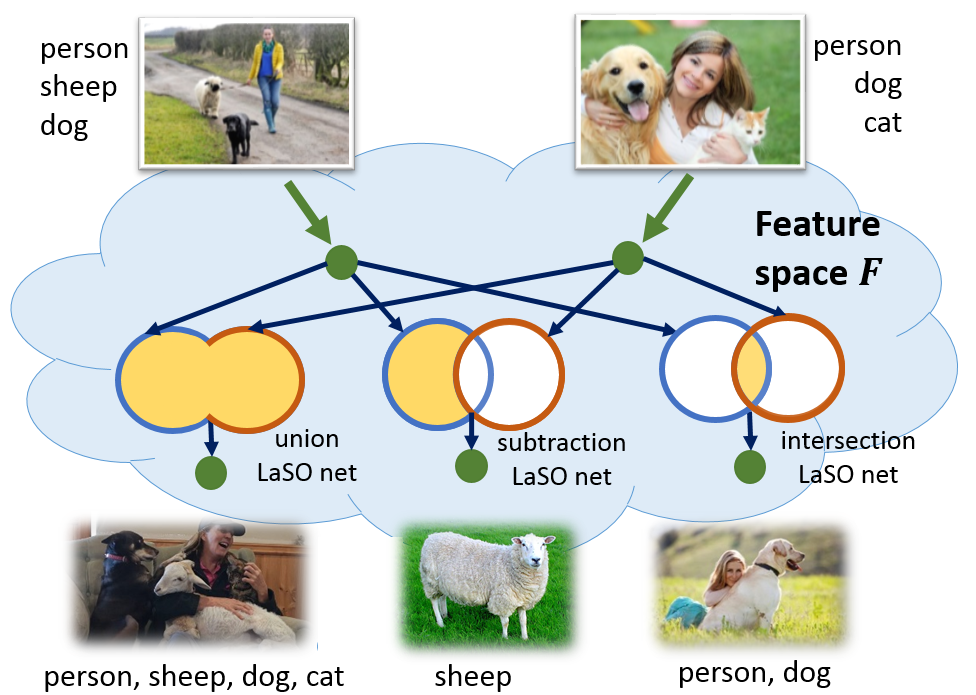}
\end{center}
   \caption{LaSO networks operating in a feature space. The goal of these networks is to synthesize new feature vectors from pairs of input vectors so that the semantic content of the synthesized vector will correspond to the prescribed operation on the source vector's label sets.}
\label{fig:SetOps}
\end{figure}

\begin{figure*}[ht]
	\centering
	\includegraphics[width=1.0\textwidth]{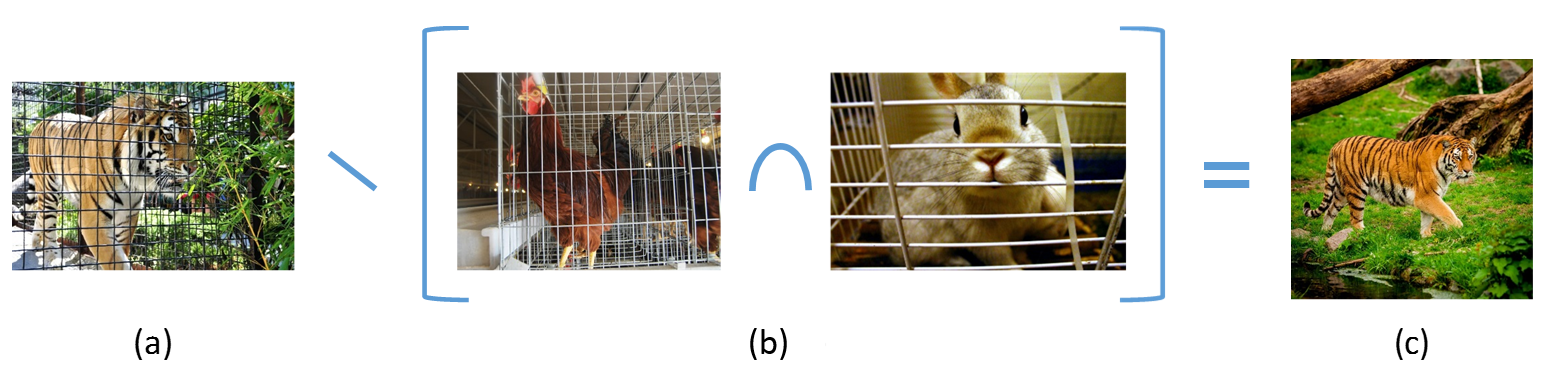}
    %  \vspace{-0.3cm}
	\caption{\textbf{LaSO concept:} manipulating the semantic content of the (small) data for better generalization to situations beyond what was originally observed. The manipulation is based on the data itself and is performed in \textit{feature space}. For real examples of our approach performing the $A \setminus \left(B \cap C\right)$ operation on real images, please see figure \ref{fig:retrieval}d.}
	\label{fig:LaSO_concept}
\end{figure*}
Deep learning excels in creating informative and discriminative feature spaces for many types of data, e.g. natural images \cite{He2017b,He2015,Krizhevsky2012}. In modern computer vision, image representation in a deep feature space is expected to encode all of the semantic content of interest, whether it is the object categories present in the image \cite{He2015}, their visual attributes \cite{Ferrari2007}, or their locations \cite{He2017b}. Usually, these feature spaces are trained using large quantities of labeled data tailored to the task \cite{Lin2014,imagenet}. 
However, in many practical applications, only a handful of examples are available for the target task; this scenario is known as  few-shot learning \cite{Vinyals2016}.
% However, in many practical applications, known as the few-shot learning, only a handful of examples are available for the target task.

In few-shot learning, the feature spaces are usually transferred from other tasks, either directly or through meta-learning that allows generating these spaces on the fly (see survey of such techniques in Section \ref{sec:related_work}). 
%Please see Section \ref{sec:related_work} for a review of the few-shot works.
One popular approach for few-shot learning is the generative one \cite{Hariharan2017,Park2015,Schwartz2018,Zhu2017}: Many new examples in the chosen feature space are generated from the few given training examples; these synthesized samples
are in turn used to improve the generalization of the few-shot task. Despite the increasing popularity of few-shot learning, all the current works on few-shot classification deal with a single (class) label per data point (e.g. $C(Img) = dog$), and not with the multi-label case (e.g. $C(Img) = \{dog,leash,person,forest\}$).

In this paper, we propose a new kind of a generative approach to few-shot learning. It explicitly targets multi-label samples; even more so, through its task definition, % \js{it can handle implicit labels, characteristic \sout{for} of training data, that restrict generalization to novel samples.}
it targets cases where the labels are not necessarily explicitly defined a-priori. 
As an illustrative example, please consider the situation depicted in Figure \ref{fig:LaSO_concept}. Suppose you wish to build a (multi-label) classifier for wild animals. You go to a zoo and take a few photos of each animal (so the learning task is a \textit{few-shot}). But alas, all of the animals are caged (Figure \ref{fig:LaSO_concept}(a)) and this few-shot trained classifier is likely to have some difficulty with the generalization to animals in the wild (Figure \ref{fig:LaSO_concept}(c)). Note that in this case, the label `caged' is not even part of the label vocabulary used for the manual annotation (here the vocabulary only contains animals). 
% This situation is an example of training \textit{dataset bias} \cite{Torralba2011}, e.g. a set of background visual concepts common to the images in the dataset, and the problem we would like to solve is one of \textit{out-of-context} recognition \cite{Rozenfeld2018,Choi2012} (here the `caged' context is inherited from the training dataset).

To address this issue, we propose having neural networks that can manipulate the `semantic content' of the samples in feature space `by example' (e.g. suppress in a feature vector elements corresponding to labels that correspond to another feature vector). For instance, consider having a model, $M_{int}$, that can accept two images with caged animals in some feature space (Figure \ref{fig:LaSO_concept}(b)), and produce a feature vector representing their common semantic content.
Since the shared (implicit) concept here is the `cage', it should end up with a feature vector representing `caged' (that is if we had a classifier for `caged' it would fire on this vector), but no longer representing either of the caged animals appearing in the original intersected images (rooster and a rabbit in this case). Then consider having another model, $M_{sub}$, that can implicitly remove concepts present in one sample from another sample (again in feature space).
We can then apply $M_{sub}$ on the caged tiger and the feature vector representing `caged' that we obtained using $M_{int}$, thus effectively getting a feature vector for a `tiger in the wild'. 
% In general, this concept may be considered for removing any dataset bias \cite{Torralba2011}. 
Please see Figure \ref{fig:retrieval}(d) for examples of our proposed approach performing $A \setminus \left(B \cap C\right)$ on real images.

Equipped with this concept, we propose to build and train a complete set of sample-based content manipulation models in feature space, namely $M_{int}$ for the label set \textit{intersection} operation, $M_{uni}$ for the label set \textit{union}, and $M_{sub}$ for the label set \textit{subtraction}. We call these models Label Set Operations networks (or LaSO nets for short).
A schematic illustration is given in Figure \ref{fig:SetOps}. The pair of images, entering the system, are converted to feature vectors using some backbone network and then processed by any of the aforementioned manipulation networks to produce feature vectors with corresponding label sets.

In Section \ref{sec:results} (results), we show that our proposed approach exhibits an ability to generalize to unseen (unlabeled) concepts allowing us to apply the LaSO nets to semantic concepts not present in the set of previously observed labels (like the label `caged' in the previous example). We show this by demonstrating a far from chance level of success of our approach manipulating labels unseen during training. This in turn allows our approach to be applied in the multi-label few-shot scenario, generating synthetic examples by manipulating novel classes unseen during training.

To summarize, our main contributions are threefold. \textbf{First}, we propose a method for the \textit{few-shot multi-label} learning task, a novel direction in few-shot learning research, not addressed so far in the literature. \textbf{Second}, we propose a novel concept of \textit{by-example label-set manipulation} in feature space, allowing the generation of new multi-label samples of interest by combining other samples. In our approach the manipulation on the labels of the combined samples is defined by the semantic content of the samples themselves and hence does not necessarily require an explicit supervised pre-training of all possible desired manipulations.
\textbf{Third}, we offer the community a new first benchmark  for the few-shot multi-label learning task, accompanied with a set of performance evaluations and baseline comparisons.
%\textbf{Third}, we show how, using the proposed approach, the multi-label few-shot classification performance can be significantly improved, compared to an elaborate choice of common practice baselines. Specifically, we achieve improvement of about 4 to 5 points on top of the best performing baseline. 

This paper is organized as follows. Section \ref{sec:related_work} reviews related work in the fields of few-shot learning and training samples augmentation. Section \ref{sec:method} explains the technical details of our proposed approach. Section \ref{sec:results} reviews the various experiments and results. Finally, Section~\ref{sec:summary} presents our conclusions and suggestions for future work.

\section{Related Work}
\label{sec:related_work}

Recently, the problem of few-shot learning has received much attention in the computer vision community. In the Meta-Learning (or learning-to-learn) approach \cite{Finn2017,Li2017,Munkhdalai2017,Ravi2017,Snell2017,Vinyals2016,Zhou2018}, classification models are trained not on individual annotated samples,
%category instances,
but rather on instances of the few-shot learning task, comprised of a small training set and a number of query samples. The goal of a meta-learning approach is to learn a model that produces models for any such few-shot task, usually without (or with only a short) fine-tuning for each task.

% the task is defined by both the training dataset and the query data, and the role of the model is to produce, based on the training data of the task, a good recognition engine for the query images. Visual categories, appearing in the task, are not present in the off-line training stage of the model, so the solution (say, a classifier) needs to be built 'on-the-fly', with no or short training. A number of works exploring this paradigm are \cite{Vinyals2016,Snell2017,Finn2017,Li2017,Zhou2018, Ravi2017,Munkhdalai2017}.

Another line of works in few-shot learning is characterized by enriching the small initial training dataset using data augmentation and data synthesis techniques. Simple image transformations (horizontal flips, scaling, shifts), have been exploited in the machine learning community from the beginning. The work in \cite{Ratner2017} takes this type of augmentation to the next level by learning a sequences of user-defined (black-box) transformations, along with their parameters, that keep the objects recognizable. 

In the synthesis approaches, new examples are generated based on the few provided labeled ones (in out-of-sample manner). Some works render synthetic examples using geometric deformations \cite{Park2015} or CNNs \cite{Dosovitskiy2017,Su2015}; specifically, a strong recent trend is to generate examples using Generative Adversarial Networks (GANs) \cite{Durugkar2017,Goodfellow2014,Huang2017,Zhu2016,Mao2016,Radford2015,Reed2018,Zhu2017}. In other works, the example synthesis is done using additional semantic information \cite{Chen2018,Yu2017}, relative linear offsets between elements of the same category in feature space \cite{Hariharan2017}, learning to extract and apply a non-linear transformation between pairs of examples of the same category \cite{Schwartz2018}, or training augmentation and classification modules end-to-end in a closed loop \cite{Wang2018}.

The approach for sample synthesis taken in this work relies on generating new samples corresponding, on the level of semantic labels, to intersection, union or subtraction of the labels present in two input samples. These labels may be objects or attributes that are present in the input samples. The set operations are non-degenerate only in the \textit{multi-label} scenario, either when each image contains multiple objects (e.g. MS-COCO dataset) or a single objects with multiple attributes (e.g., CelebA dataset).

Some prior works on multi-label classification improve upon the straightforward approach of having an independent classifier per label by learning label correlations within images (see \cite{Wang2016b} for an extensive review). Yet, in the few-shot domain, this information cannot be exploited for a new task, which contains unseen categories. In \cite{Rios2018}, the task of few-shot multi-label text classification is addressed, relying on the structure of the label space specific to text. To the best of our knowledge, there is no prior work of multi-label few-shot visual categories classification.

In the domain of object composition, \cite{Nagarajan2018} models attributes as operators, learning a semantic embedding that explicitly factors out attributes from their accompanying objects, in order to recognize unseen attribute-object compositions. In \cite{Azadi2018},
%In the domain of object composition, a somewhat related problem is addressed  \cite{Azadi2018}. 
a pipeline for integrating two visual objects is proposed, for the purpose of generating images composed of the two objects, spatially combined (tested on synthetic data). This task is very different than the one we would like to address, as: (1) a spatial combination of objects requires to learn occlusions; and (2) the composition takes place in the image space, rather than on the features level, which we aim at. The latter provides the ability to use existing feature extractors (such as Inception \cite{Szegedy2015} or ResNet \cite{He2015}) more easily, which makes it much more applicable, e.g. to few-shot classification.

\section{Method}
\label{sec:method}
\begin{figure*}[ht]
	\centering
	\includegraphics[width=1.0\textwidth,trim=0cm 2cm 0cm 0cm,clip]{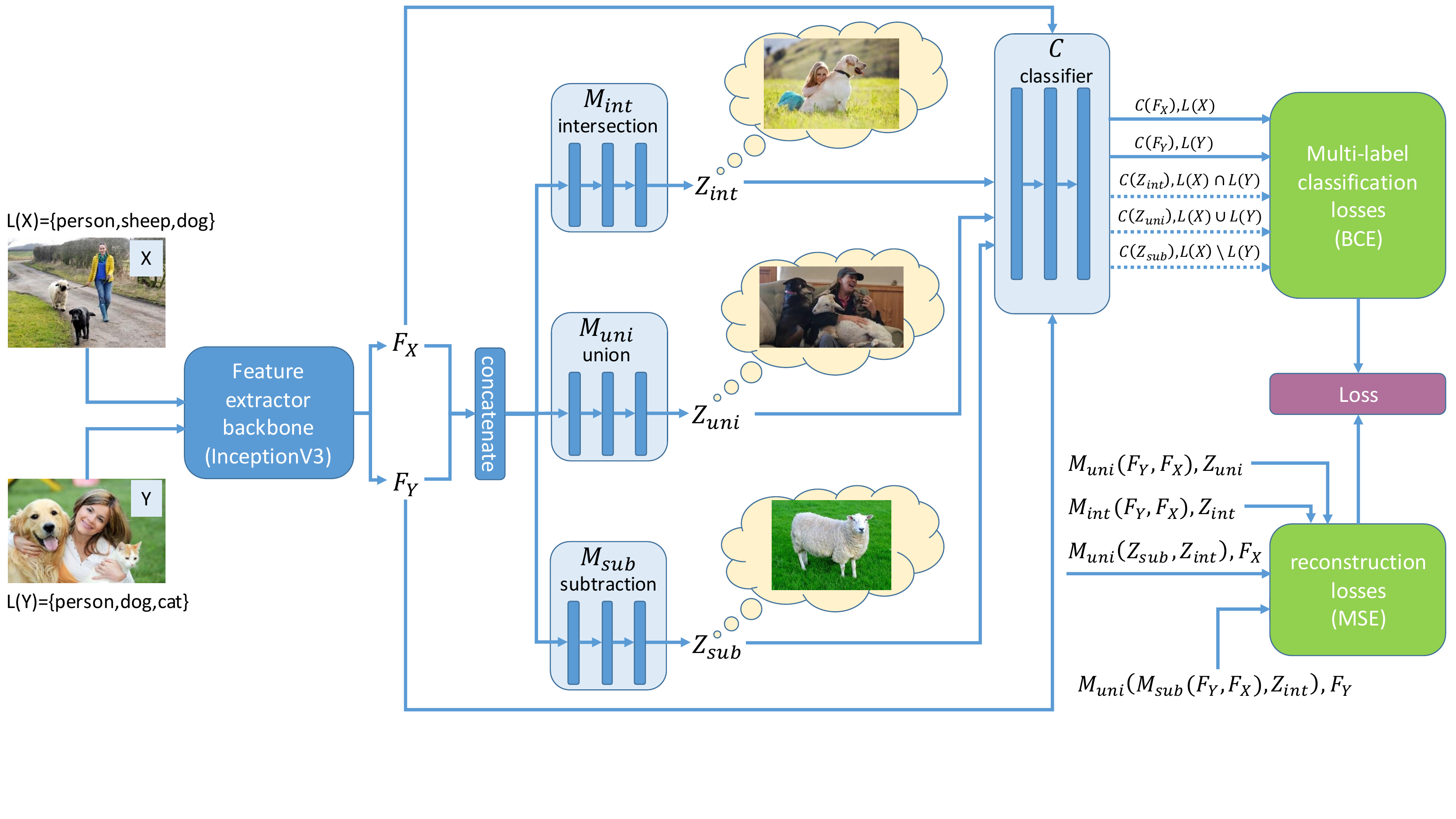}
    %  \vspace{-0.3cm}
	\caption{\textbf{LaSO model:} schematic illustration of all the components of the proposed approach (including training losses). }
	\label{fig:LaSO_model}
\end{figure*}

Our approach is schematically illustrated in Figure \ref{fig:LaSO_model}. Input images $X$ and $Y$, each with a corresponding set of multiple labels, $L(X), L(Y) \subseteq	\mathcal{L}$ respectively, are represented in the joint feature space $\mathcal{F}$ as $F_X$ and $F_Y$. This space $\mathcal{F}$ is realized using a backbone feature extractor network $\mathcal{B}$; we have used InceptionV3 \cite{Szegedy2015} and ResNet-34 \cite{He2015} backbones in our experiments. Three LaSO networks $M_{int}$, $M_{uni}$, and $M_{sub}$ receive the concatenated $F_X$ and $F_Y$ and are trained to synthesize feature vectors in the same space $\mathcal{F}$. As the name (\textit{int=intersection}) suggests, $M_{int}$'s goal is to synthesize a feature vector
\begin{equation}
    M_{int}\left( F_X, F_Y \right) = Z_{int} \in \mathcal{F},
\end{equation}
which corresponds to a hypothetical image $I$, such that $\mathcal{B}(I) = Z_{int}$ and $L(I)=L(X) \cap L(Y)$.
In other words, this means that if a human would observe and label $I$, it would receive $L(X) \cap L(Y)$ as its label set. 
Similarly, $M_{uni}$ and $M_{sub}$ output $Z_{uni},Z_{sub} \in \mathcal{F}$ that are expected to correspond to the union of the label sets $L(X) \cup L(Y)$, and the subtraction of the label sets $L(X) \setminus L(Y)$ respectively. 

Note that although we use a pre-defined set of labels $\mathcal{L}$ for training our models, we can expect that during training, the networks will also generalize to labels which are not part of $\mathcal{L}$. 
This is possible because LaSO nets receive no explicit label information as input (neither during training, nor during use). They are forced to learn to synthesize vectors corresponding to the desired label sets implicitly, only by observing $F_X$ and $F_Y$ as their inputs, without being explicitly given their labels.
% , and are guided only by the classification and reconstruction losses gradients computed based on the LaSO nets outputs.
%This is possible because the label set manipulation using LaSO networks is controlled only by the inputs to the networks and not via any explicit supervisedly trained external control based on explicit labels. 
%
In Section \ref{sec:results} (Results) we test this ability of our networks to generalize to novel categories.

The source feature vectors, $F_X$ and $F_Y$, and the outputs of the LaSO networks, namely $Z_{int}$, $Z_{uni}$, and $Z_{sub}$, are fed into a classifier $C$. We use the Binary Cross-Entropy  ($BCE$, aka Sigmoid-Cross-Entropy) multi-label classification loss in order to train $C$ and the LaSO networks:
\begin{equation}
BCE(s,l) = -\sum_i l_i\log\sigma(s_i)+(1-l_i)\log(1-\sigma(s_i)),
\end{equation}
with the sigmoid $\sigma(x)=(1+\exp(x))^{-1}$, the vector $s$ being the classifier scores, $l$ being the desired (binary) labels vector, and $i$ the class indices. To train the classifier $C$ we use only the combination of the losses from the source feature vectors:
\begin{equation}
C_{loss}=BCE(C(F_X),L(X))+BCE(C(F_Y),L(Y)),
\end{equation}
where $C(\cdot)$ stands for the classifier $C$ output score vector. The LaSO networks are trained using:
\begin{align}
LaSO_{loss}=&BCE(C(Z_{int}),L(X) \cap L(Y))+\\
&BCE(C(Z_{uni}),L(X) \cup L(Y))+ \nonumber\\
&BCE(C(Z_{sub}),L(X) \setminus L(Y)) \nonumber
\end{align}
For the LaSO updates the classifier $C$ is kept fixed and only used for passing gradients backwards. Note that the used losses decouple the training of $C$ and the LaSO networks.

In addition, our model includes a set of Mean Square Error (MSE) based reconstruction losses. The first loss is used to enforce symmetry for the symmetric \textit{intersection} and \textit{union} operations.
This loss $R^{sym}_{loss}$, is realized as the MSE between $Z_{int}=M_{int}\left(F_X,F_Y\right)$, $Z_{uni}=M_{uni}\left(F_X,F_Y\right)$ and the vectors obtained from the corresponding networks with the reversed order of the inputs:
\begin{align}
R^{sym}_{loss}=&\frac{1}{n}\|Z_{int} - M_{int}\left(F_Y,F_X\right)\|_2 + \\
&\frac{1}{n}\|Z_{uni} - M_{uni}\left(F_Y,F_X\right)\|_2 \nonumber
\end{align}
Please note that $M_{int}(F_X,F_Y)$ and $M_{int}(F_Y,F_X)$ invoke the same instance of $M_{int}$. Same holds for any LaSO network that is invoked multiple times in our construction.

The second loss is used in order to reduce the chance of mode collapse that could cause a semi-fixed output for each possible label set combination. For example, in case of a mode collapse, we could observe very similar outputs of the network $M_{int}$ for many different pairs of images with the same set of shared labels.
The mode collapse related reconstruction loss, $R^{mc}_{loss}$, is realized as the MSE loss between $F_X$, $F_Y$ and the outputs of simple expressions (generated by some combinations of the LaSO networks) that produce feature vectors that should correspond to the original label sets $L(X)$ and $L(Y)$ by set-theoretic considerations:
\begin{align}
R^{mc}_{loss}=&\frac{1}{n}\|F_X - M_{uni}\left(Z_{sub},Z_{int}\right)\|_2^2 + \\
&\frac{1}{n}\|F_Y - M_{uni}\left(M_{sub}(F_Y,F_X),Z_{int}\right)\|_2^2, \nonumber
\end{align}
where $n$ is the length of $F_X$.

\begin{figure*}[ht]
	\centering
	\includegraphics[width=1.0\textwidth,trim=0cm 0cm 0cm 0cm,clip]{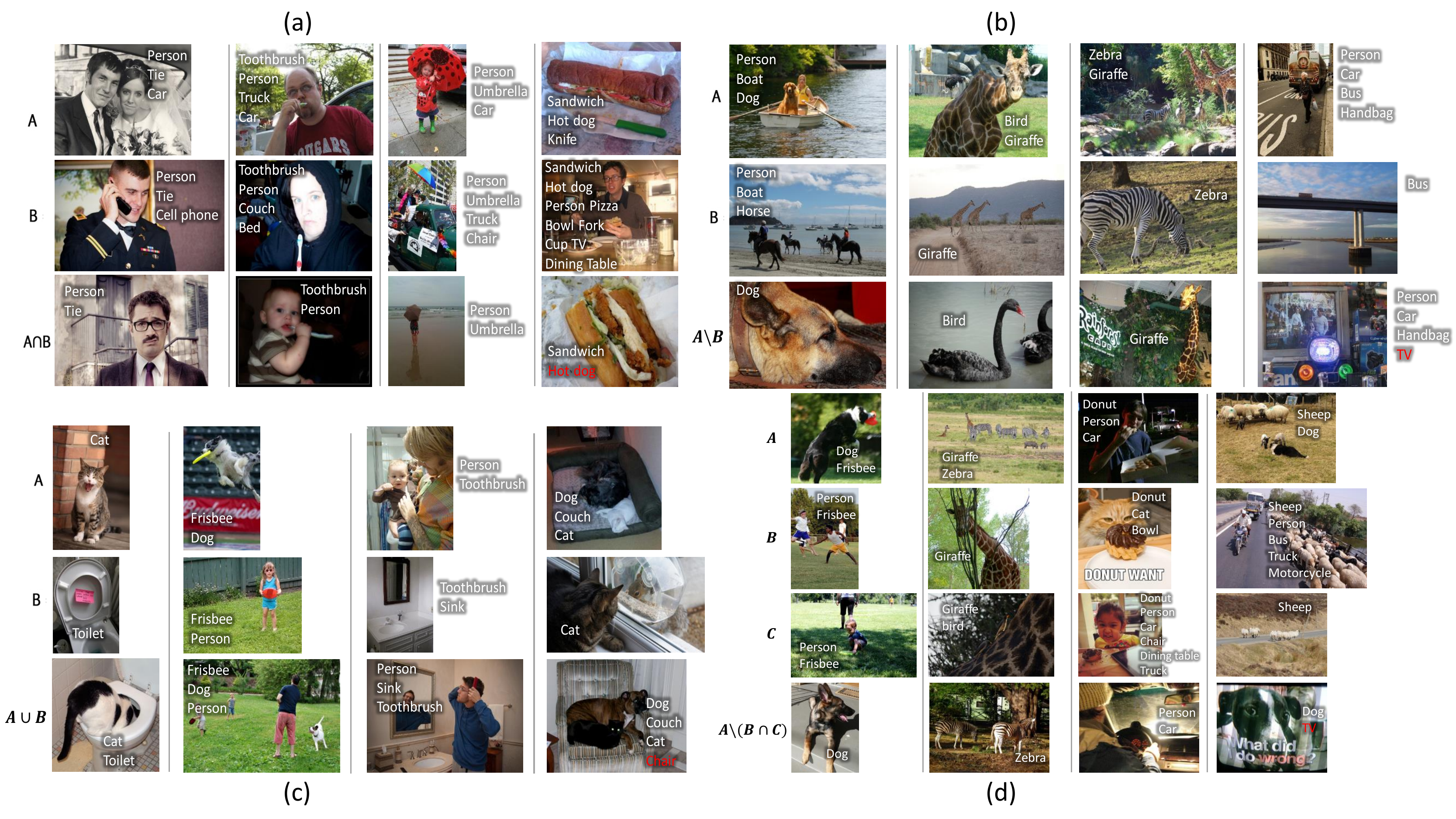}
    %  \vspace{-0.3cm}
	\caption{\textbf{Testing LaSO networks using retrieval:} $A$ and $B$ feature vectors are inputs to LaSO nets and the nearest neighbor image in feature space to the output feature vector is shown below each pair. For each operation we show three successful examples and one failure case highlighting the errornous label in red. Best viewed in color. (a) intersection retrieval examples; (b) subtraction retrieval examples; (c) union retrieval examples; (d) $A \setminus B \cap C$ retrieval examples.}
	\label{fig:retrieval}
\end{figure*}

\subsection{Implementation details}
We have implemented our approach using PyTorch 1.0 \cite{Paszke2017}. The InceptionV3 and the ResNet-34 feature extractor backbones are pre-trained from scratch using the corresponding training sets as described in Section \ref{sec:results} (Results). The LaSO networks are implemented as Multi-Layer Perceptrons (MLPs) consisting of 3 or 4 blocks. Each block contains a fully-connected layer followed by batch-normalization, leaky-ReLU, and dropout. A future work may explore additional architectures for the LaSO nets, e.g. encoder-decoder and residual based architectures. During training, we used batch size of 16, initial learning rate of 0.001, learning rate reduced on loss plateau with factor 0.3. The optimization is performed with the Adam optimizer \cite{Kingma2015} with parameters $(0.9,0.999)$.
	
\section{Results}
\label{sec:results}

An image usually contains multiple object instances that can be translated to a set of unique category labels. Object detection and segmentation datasets are a great source of multi-object labels. Indeed, by throwing away the bounding boxes and segmentation masks, and keeping only the unique category labels set we can transform any such dataset into a multi-label classification one. In our experiments we used the popular (and challenging) MS-COCO \cite{Lin2014} dataset as the source of multi-object labels.

An object, e.g. a face, can be described in terms of its various attribute labels. To test our approach on the task of manipulating the attribute-based multi-label data, we have used the CelebA \cite{Liu2015} dataset. In CelebA experiments we have used its 40 facial attribute annotations as labels.

\subsection{MS-COCO experiments}
\label{sec:results-coco}

For MS-COCO experiments we have used the COCO 2014 train and validation sets. The 80 COCO categories were randomly split into 64 `seen' and 16 `unseen' categories. The unseen categories were: \textit{bicycle, boat, stop sign, bird, backpack, frisbee, snowboard, surfboard, cup, fork, spoon, broccoli, chair, keyboard, microwave}, and \textit{vase}. We filtered the COCO train set leaving only images that did not contain any of the 16 unseen category labels and used this filtered set to train our feature extractor backbone (InceptionV3) and the LaSO models (as described in section \ref{sec:method}). Before training jointly with the LaSO models, the feature extractor backbone was first pre-trained separately as a multi-label classifier for the 64 seen categories on the filtered training set using the standard BCE classification loss.

\begin{table}[h]
	\small
	\centering
	\begin{tabular}{lcc}
		\toprule 
		& $64$ seen classes & $16$ unseen classes \\
		\midrule
		intersection  & 77 & 48 \\
		union & 80 & 61 \\
		subtraction & 43 & 14 \\
		\midrule
		original (non-manipulated) \\feature vectors  & 75 & 79 \\
		\bottomrule
	\end{tabular}		
	\caption{Evaluating feature vectors synthesized by the LaSO networks using the \textit{classification} performance on the $64$ \textit{seen} and on the $16$ \textit{unseen} MS-COCO categories. Classification is performed w.r.t. the expected label set after each type of operation, and on the original feature vectors for reference. All tests are performed on the MS-COCO validation set, not used for training. Numbers are in mAP $\%$. }
	\label{tab:coco-classification}
\end{table}

\begin{table*}[ht]
	\small
	\centering
	\begin{tabular}{lcccccc}
		\toprule 
		& \multicolumn{3}{c}{$64$ seen classes} & \multicolumn{3}{c}{$16$ unseen classes} \\
		\cmidrule(lr){2-4}\cmidrule(lr){5-7}
		  & top-1 & top-3 & top-5 & top-1 & top-3 & top-5 \\
		\midrule
		intersection  & 0.7 & 0.79 & 0.82 & 0.47 & 0.71 & 0.78 \\
		union & 0.61 & 0.71 & 0.74 & 0.44 & 0.64 & 0.71 \\
		subtraction & 0.19 & 0.32 & 0.4 & 0.21 & 0.4 & 0.51 \\
		\midrule
		original (non-manipulated) \\feature vectors  & 0.56 & 0.72 & 0.76 &  0.56 & 0.75 & 0.81 \\
		\bottomrule
	\end{tabular}		
	\caption{Evaluating feature vectors synthesized by the LaSO networks using the \textit{retrieval} performance on the $64$ \textit{seen} and on the $16$ \textit{unseen} MS-COCO categories (Sec.~\ref{eval_label_manipulation}). Retrieval quality is measured w.r.t. the expected label set after each type of operation. All tests are performed on the MS-COCO validation set, not used for training. Numbers are \textit{mean Intersection over Union} (mIoU) between the label sets of the retrieved samples and the expected label set, the mean is taken over the different queries. The top-$k$ averages the maximum IoU obtained among closest $k$ retrieved samples. In order to assess the expected range of retrieval performance in feature space $\mathcal{F}$, we also provide a reference of the same quality measurement for retrieval using the the original non-manipulated feature vectors. }
	\label{tab:coco-retreival}
\end{table*}

\subsubsection{Evaluating the label set manipulation capability of the LaSO networks}
\label{eval_label_manipulation}
%synthesis
We used the COCO validation set to test the performance of the resulting LaSO models for the label set intersection, union, and subtraction operations. We applied two methods for this evaluation, one using classification and the other using retrieval.

In the classification tests we have used a classifier pre-trained on the feature space $\mathcal{F}$ (generated by the backbone feature extractor model) to test the LaSO networks. To this end, we have randomly paired all of the validation set images and tested each LaSO operation network on each pair. For any pair of images $X$ and $Y$, and their corresponding feature vectors $F_X$ and $F_Y$, the outcome of $M_o(F_X,F_Y)$, where $o \in \{uni,int,sub\}$, was fed to the classifier and its resulting class scores were evaluated vs the expected label-set resulting from applying the set operation $o$ on $L(X)$ and $L(Y)$. We performed two separate evaluations, one for the seen and the other for the unseen categories. In each of the tests we compute the Average Precision (AP) for each category and report the mean AP (mAP) computed over the categories in each (seen / unseen) set. 
% To compute the AP for a specific category, the collection of scores, produced by the classifier for that specific category, is obtained for the set of test images, along with the ground truth binary labels, indicating whether this category was truly present in the image. This data is then used to compute the AP using the standard method of computing the area under the precision-recall curve. 
%(LK: someone might claim they are insulted by explaining simple things... idk)

For the seen categories we used the classifier that was obtained when the backbone $\mathcal{B}$ model was pre-trained as a classifier on the 64 seen categories set. For the 16 unseen categories, the 16-way classifier, used for the evaluation, was pre-trained on the images of the COCO training set containing instances of these 16 categories. 
%This classifier for unseen categories was trained over the
For its training, we used the same feature space $\mathcal{F}$ generated by our backbone $\mathcal{B}$. The reason is that the trained LaSO networks can only operate in this space. The results of the classification based evaluation experiments are summarized in Table~\ref{tab:coco-classification}. 
%From the obtained results we can see 
On the set of seen categories, for the union and intersection operations, the LaSO networks managed to learn to synthesize feature vectors which through the eyes of the classifier are seen as comparable (even slightly better) to the original non-manipulated feature vectors. On the unseen categories there is still room for improvement. Yet even there the results are well above chance, indicating that despite not observing any of the unseen categories during training, the LaSO label set manipulation operations managed to generalize beyond the original training labels. This opens the door for the multi-label few-shot experiments on the set of the unseen categories presented in section~\ref{sec:multi-label-few-shot} below.

In the retrieval tests we have evaluated the synthesized feature vectors directly without using any classifier. We used nearest neighbor search in a large pool of feature vectors of real images with ground truth labels. To this end, as in the classification tests, validation images were randomly paired and passed through the LaSO networks resulting in synthesized feature vectors with an expected set of labels (according to the operation). The synthesized feature vectors were then used to retrieve the first $k$ nearest neighbors (NNs) in the validation set. Please see Fig.~\ref{fig:retrieval} for some examples of inputs to different LaSO nets, and the corresponding retrieved NNs. For each of the resulting NNs, Intersection over Union (IoU) was computed between the ground truth label-set of the NN and the expected label-set of the synthesized vector. Then maximum IoU was computed on the top-$k$ NNs. In Table~\ref{tab:coco-retreival} we report average IoU computed over the entire set of the synthesized vectors, for different $k \in \{1,3,5\}$ and for the seen and unseen sets of categories separately. For reference, we also repeat the retrieval performance evaluation as above for the original non-manipulated feature vectors in order to set a frame of reference. Again, as can be seen from the results, in terms of retrieval, the feature vectors synthesized by the LaSO networks for the intersection and the union operations are performing on par with the original non-manipulated ones. The performance is slightly better for some of the $k$ on the set of seen categories, and quite close on the unseen ones. This again provides evidence for the ability of the LaSO networks to generalize to unseen categories
%underlines the generalization obtained by the LaSO networks in training 
and supports their use for performing augmentation synthesis for few-shot multi-label training (Sec.~\ref{sec:multi-label-few-shot}).

\subsubsection{Analytic approximations to set operations}
\label{sec:results-ablation}

Using the (naive) interpretation of the feature vectors in the space $\mathcal{F}$ as collections of individual features correlated with the appearance of specific visual labels, we can consider analytic operations on pairs of feature vectors which mimic the effects of the set operations in the label space. This enables a simpler version of our method, which does not involve learned LaSO networks, but still generates synthetic features that can contribute to multi-label few-shot classifier training as will be demonstrated in section \ref{sec:multi-label-few-shot}.

Denoting the input to LaSO networks by $F_X, F_Y \in \mathcal{F}$, as in the Fig.~\ref{fig:LaSO_model}, we have defined and evaluated the following set of analytic LaSO alternatives:

\begin{center}
    \small
	\begin{tabular}{lcc}
	\toprule 
	Operator & Expression 1 & Expression 2 \\
	\midrule 
    Union  & $F_X + F_Y$   & $\max(F_X,F_Y)$ \\
    Intersection & $F_X \cdot F_Y$   & $\min(F_X,F_Y)$ \\
    Subtraction & $F_X - F_Y$  & ReLU$(F_X-F_Y)$ \\
    \bottomrule 
    \end{tabular}
\end{center}

We defined this set of alternatives drawing intuition from the DCGAN paper \cite{Radford2015}, that has proposed GAN arithmetics as an interesting possibility of manipulating images in the space of GAN random seeds. In our case, we are not assuming a (well) trained GAN for our multi-label data, and explore a simpler variant, directly manipulating feature vectors in $\mathcal{F}$. Table \ref{tab:ablation} summarizes the comparison between the top performing analytic and the learned LaSO variants on both the COCO and the CelebA datasets. In both experiments, the top performing analytic expressions were $\max(F_X,F_Y)$ for the union, $\min(F_X,F_Y)$ for the intersection, and $ReLU(F_X-F_Y)$ for the subtraction. As can be seen, the learned LaSO networks outperform the simpler analytic alternatives in almost all cases, yet in some cases the analytic versions are not far behind,
%perform close
indicating them as additional good candidates for being used for augmentation synthesis in few-shot multi-label experiments in section \ref{sec:multi-label-few-shot}.

\begin{table}[ht]
  \small \centering
  \begin{tabular}{lcccc}
    \toprule
     dataset & method  & subtraction & intersection & union \\
    \midrule
    \multirowcell{2}[0pt][l]{\textbf{MS-COCO}} 
     & analytic & 29.0 & 74.7 & 76.5 \\
     & learned & \textbf{43.0} & \textbf{77.0} & \textbf{80.0}  \\
    \midrule
    \multirowcell{2}[0pt][l]{\textbf{CelebA}} 
     & analytic & 37.0 & \textbf{52.0} & 47.0 \\
     & learned & \textbf{69.0} & 48.0 & \textbf{75}  \\
    \bottomrule
  \end{tabular}
  \caption{\textbf{Ablation study:} comparing the learned operators with analytic alternatives. All numbers are in mAP $\%$.}
  \label{tab:ablation}
\end{table}

\subsubsection{Multi-label few-shot classification experiments}
\label{sec:multi-label-few-shot}

In this section we explore an interesting application of the label-set manipulation concept - serving as a (learned) augmentation synthesis method for training a multi-label few-shot classifier.
As opposed to the well-studied single-label few-shot classification, in the multi-label few-shot scenario the examples of different categories are only provided in groups. This renders the existing techniques for few-shot classification inapplicable, and to the best of our knowledge, this problem was not addressed before.
%As opposed to the more classical, single label per image, few-shot classification problem explored by many works, in the multi-label few shot case examples of different label categories are not provided separately, but almost always come in groups. This makes the few annotated multi-label examples a superposition of signals from different categories of interest, and rendering the existing single-label few-shot approaches inapplicable. To the best of our knowledge, as we write this, there are no existing techniques or benchmarks for multi-label few-shot classification. 

Therefore, we propose our own benchmark and a first set of results for this problem, comparing our approach to multiple natural baselines. The baselines are: (A) training directly on the small labeled set, (B) using standard (basic) image augmentation while training on the small labeled set, and (C) using the mixUp \cite{Zhang2017} augmentation technique.
%more advanced augmentation techniques like mixUp \cite{Zhang2017}.As instances of our proposed label set manipulation approach, we are considering both the learned LaSO networks as well as the analytical alternatives discussed in section \ref{sec:results-ablation}. 
We compared these baselines to both the learned LaSO networks and the analytical alternatives discussed in Section \ref{sec:results-ablation}.
\begin{table}[ht]
	\small
	\centering
	\begin{tabular}{lcc}
		\toprule
    & 1-shot & 5-shot \\
		\midrule
		B1: no augmentation  & $39.2$ & $49.4$ \\
		B2: basic aug.  & $39.2$ & $52.7$ \\
		B3: mixUP aug.  & $40.2$ & $54.0$ \\
		\midrule
% 		analytic subtraction aug. & $X$ & $X$ \\
		analytic intersection  aug. & $40.7$ & $55.4$ \\
		analytic union aug.  & $44.5$ & $55.6$ \\
		\midrule
% 		learned subtraction aug. & $40.0$ & $54.1$ \\
		learned intersection aug. & $40.5$ & $57.2$ \\
		learned union aug. & $\textbf{45.3}$ & $\textbf{58.1}$ \\
		%
% 		B1: no augmentation  & $39.2 \pm 5$ & $49.4 \pm 4$ \\
% 		B2: basic aug.  & $39.2 \pm 6$ & $52.7 \pm 3$ \\
% 		B3: mixUP aug.  & $40.2 \pm 4$ & $54 \pm 3$ \\
% 		\midrule
% 		fixed intersection  aug. & $40.7 \pm 5$ & $55.4 \pm 4$ \\
% 		fixed union aug.  & $44.5 \pm 4$ & $55.6 \pm 3$ \\
% 		\midrule
% 		learned intersection aug. & $40.5 \pm 3$ & $57.2 \pm 3$ \\
% 		learned union aug. & $\textbf{45.3} \pm \textbf{4}$ & $\textbf{58.1} \pm \textbf{3}$ \\
% 		\midrule
% 		fully trained \\ (non few-shot)& \multicolumn{2}{c}{79.0} \\
		\bottomrule		
	\end{tabular}		
	\caption{ Multi-label few-shot mAP (in $\%$) on 16 unseen categories from MS-COCO. The feature extractor and the LaSO networks are trained on the remaining 64 MS-COCO categories. Average of 10 runs are reported, tested on the entire MS-COCO test set. MixUP baseline uses the original code of \cite{Zhang2017}. 
% 	\rg{RG: The fully trained makes all the other results seem very bad. Why we need it here? Also, can we combine intersection and union together or all the three together (including subtraction)? I assume subtraction did not help but we need to say something about why we do not use it here. Another thing is that one may wonder what happens when we compare to mix-up in the case where we have lots of examples. Can the set operations improve network training also when have more data like mixup does?}
    }
	\label{tab:aug_test}
\end{table}

As our benchmark, we propose the set of the 16 COCO categories unseen during training. We generate 10 random episodes (few-shot train set selection) for each of the 1-shot (1 example per category) and 5-shot (5 examples per category) scenarios. The same episodes are used for all the methods: the LaSO variants and all the baselines. During episode construction we maintained a histogram of the label counts ensuring that a total of 1 example per category appears in the episode for 1-shot scenario and 5 examples in 5-shot scenario respectively. Of course due to the random nature of the episodes, this balancing is not always possible, and hence in some episodes the amount of labels per category could exceed 1 or 5 (just by 1 in the majority of the cases). But since same exact episodes are used for all the compared approaches the comparison are fair. The entire COCO validation set (considering only the 16 unseen categories annotations) is used for testing the classifiers trained on each of the episodes. 

All the training and the validation images were converted to the same feature space $\mathcal{F}$ created by our feature extraction backbone, the training and the augmentation were performed on top of $\mathcal{F}$ (except for the standard augmentation that was applied to the images and then converted to $\mathcal{F}$ by the backbone). Random pairs of examples from the small ($1$ or $5$-shot $\times$ 16 categories) training set were used for label-set manipulations. For all the augmentation baselines and all variants of our method, same number of samples were synthesized per training epoch. On all compared approaches the classifiers trained on each of the episodes were trained using 40 SGD epochs (as we experimentally verified, all of them converged before 40 epochs).

The results of this experiment are reported in Table~\ref{tab:aug_test}. All results are reported in mAP $\%$ computed over the 16 unseen categories in the entire COCO validation set. As can be seen from the results, for both 1 and 5 shot scenarios label set manipulation obtains stable gains of $5.1$ and $4.1$ mAP points respectively. This 
points towards the ability of the LaSO networks to generalize to unseen labels, also showing 
% again the generalization ability of the LaSO networks towards operating on labels unseen during training, as well as showing 
the general utility of our label-set manipulation approach in learning to augment data for training multi-label few-shot classifiers in a challenging realistic scenario (COCO).

\subsection{CelebA experiments}
\label{sec:results-celeba}

% In order to test our approach on a different 
% %major source of multi-label
% \js{kind of multi-label data}, namely object attributes, we have used the CelebA dataset \cite{Liu2015} and its 40 facial attributes.

We used the CelebA dataset \cite{Liu2015} in order to test our approach on a different kind of multi-label data, namely object attributes. The CelebA dataset contains $\sim 200$K images labeled according to 40 facial attributes.
 We pre-trained the feature extractor backbone (based on the ResNet-34) as a multi-label classifier on the training samples of the CelebA dataset. Then we trained $M_{uni}$, $M_{int}$ and $M_{sub}$ to perform the corresponding set-operations on the attribute-based multi-labels on the same training data. 
%  The LaSO networks were trained using the training samples of the CelebA dataset. 
%  The backbone and classifier networks were not modified during the training of the LaSO models. 
 We then repeated the classification based evaluation experiments and ablation studies as described for COCO in section \ref{sec:results-coco}. The test samples of the CelebA dataset were used to evaluate the performance.
 The results of the classification based evaluation are summarized in Table~\ref{tab:celeba-classification} in mAP~$\%$ computed over the 40 attributes of CelebA.
The union and subtraction LaSO networks achieve relatively high mAP while the intersection network scores lower. This can be attributed to the fact that the intersection network training is unbalanced and biased toward negative attributes (the intersection operation leaves most attributes turned off), while the precision computation is more affected by the ability to accurately predict the positive labels. Results of the ablation studies are given in Table \ref{tab:ablation}.
 
\begin{table}[ht]
	\small
	\centering
	\begin{tabular}{lc}
		\toprule 
		& $40$ facial attributes \\
		\midrule
		intersection  & 48 \\
		union & 75 \\
		subtraction & 69 \\
		\midrule
		original (non-manipulated) \\feature vectors  & 79 \\
		\bottomrule
	\end{tabular}		
	\caption{Evaluating feature vectors synthesized by the LaSO networks using the \textit{classification} performance on the $40$ facial attributes in CelebA. Classification is performed w.r.t. the expected label set after each type of operation, and on the original feature vectors for reference. All tests are performed on the CelebA test set, not used for training. Numbers are in mAP~$\%$.  }
	\label{tab:celeba-classification}
\end{table}

% \subsection{Analytic approximations to set operations}
% Using the (naive) interpretation of the feature vectors in the space $\mathcal{F}$ as collections of individual features correlated with the appearance of specific visual objects, we propose analytical operations on pairs of feature vectors which mimic the effects of the set operations in the label space. This enables us to suggest a simpler version of our method, which does not involve learned LaSO networks, and still generates synthetic features that contribute to few-shot classifier training.

% Intuitively, taking an intersection of labels in two images corresponds to selecting the features that were dominant in representations of both images in the provided input pair. Therefore, we propose to approximate the intersection operation with the elementwise minimum operator on the pair of corresponding feature vectors: 
% \begin{equation}
%   F_{X\cap Y} \sim \min(F_X,F_Y)
% \end{equation}

% dot dot dot

\section{Summary \& Conclusions}
\label{sec:summary}

In this paper we have presented the label set manipulation concept and have demonstrated its utility for a new and challenging task of the multi-label few-shot classification. Our results show that label set manipulation holds a good potential for this and potentially other interesting applications, and we hope that this paper will convince more researchers to look into this interesting problem.

Natural images are inherently multi-label. We have focused on two major sources of labels: objects and attributes. Yet, other possible sources of image labels, such as the background context, object actions, interactions and relations, etc., may be further explored in a future work.

One of the interesting future directions of this work include exploring additional architectures for the proposed LaSO networks. For example an encoder-decoder architecture, where the encoder and the decoder subnets are shared between the LaSO networks, and the label-set operations themselves are implemented between the encoder and the decoder via the analytic expressions proposed in section~\ref{sec:results-ablation}. This alternative architecture has the potential to disentangle the feature space into a basis of independent constituents related to independent labels facilitating the easier use of analytic variants in such a disentangled space.
Another interesting future research direction is to use the proposed techniques in the context of few-shot multi-label semi-supervised learning, where a large scale unlabeled data is available, and the proposed approach could be used for automatic retrieval of more auto-labeled examples with arbitrarily mixed label sets (obtained by mixing the few provided examples).
In addition, the proposed approach might also prove useful for the interesting visual dialog use case, where the user can manipulate the returned query results by pointing out or showing visual examples of what she/he likes or doesn't like.

Finally, the approach proposed in this work is related to a well known issue in Machine Learning, known as \textit{dataset bias} \cite{Torralba2011} or \textit{out-of-context} recognition \cite{Rozenfeld2018,Choi2012}. An interesting future work direction for our proposed approach is to help reducing the bias dictated by the specific provided set of images by enabling a better control over the content of the samples.

{\small
\bibliographystyle{ieee}
\bibliography{arxiv_LaSO}
}

\end{document}

%%%%%%%%% TITLE
\title{LaSO: Label-Set Operations networks for multi-label few-shot learning\\Supplementary Material}

% \and

\maketitle

Here we present some more visual examples of LaSO networks results. We apply the LaSO networks trained on 64 MS-COCO categories to images from MS-COCO validation set (unseen during training). As in Figure 4 of the paper, we use retrieval to visualize the synthesized feature vectors as the image with the closest feature vector in the validation set. We demonstrate results obtained using the learned LaSO operation, the analytic LaSO variant, and also (for reference) retrieval using the inputs to the operation, self excluded. We demonstrate example results for using LaSO to perform the label set intersection $A \cap B$ (Figure \ref{fig:intersection}), union $A \cup B$ (Figure \ref{fig:union}), subtraction $A \setminus B$ (Figure \ref{fig:subtraction}), and hybrid $A \setminus \left(B \cap C \right)$ (Figure \ref{fig:hybrid}) operations.

\begin{figure}[ht]
\begin{center}
   \includegraphics[width=0.95\textwidth]{figures/intersection_titles.pdf}
\end{center}
   \caption{Examples of label set intersection $A \cap B$ operation results using LaSO and alternatives. Learned LaSO result is framed in cyan box. True labels of the images are shown above them. `Retrieval $A$' and `retrieval $B$' are provided for reference and depict the nearest neighbors to the corresponding operation input images in feature space.}
\label{fig:intersection}
\end{figure}

\begin{figure}[ht]
\begin{center}
   \includegraphics[width=0.95\textwidth]{figures/union_titles.pdf}
\end{center}
   \caption{Examples of label set union $A \cup B$ operation results using LaSO and alternatives. Learned LaSO result is framed in cyan box. True labels of the images are shown above them. `Retrieval $A$' and `retrieval $B$' are provided for reference and depict the nearest neighbors to the corresponding operation input images in feature space.}
\label{fig:union}
\end{figure}

\begin{figure}[ht]
\begin{center}
   \includegraphics[width=0.95\textwidth]{figures/subtraction_titles.pdf}
\end{center}
   \caption{Examples of label set subtraction $A \setminus B$ operation results using LaSO and alternatives. Learned LaSO result is framed in cyan box. True labels of the images are shown above them. `Retrieval $A$' and `retrieval $B$' are provided for reference and depict the nearest neighbors to the corresponding operation input images in feature space.}
\label{fig:subtraction}
\end{figure}

\begin{figure}[ht]
\begin{center}
   \includegraphics[width=\textwidth]{figures/combined_titles.pdf}
\end{center}
   \caption{Examples of label set hybrid $A \setminus \left(B \cap C \right)$ operation results using LaSO and alternatives. Learned LaSO result is framed in cyan box. True labels of the images are shown above them. `Retrieval $A$', `retrieval $B$', and 'retrieval $C$' are provided for reference and depict the nearest neighbors to the corresponding operation input images in feature space.}
\label{fig:hybrid}
\end{figure}